\documentclass[10pt,letterpaper]{article}
\usepackage[top=0.85in,left=1.75in,footskip=0.75in,marginparwidth=2in]{geometry}
\usepackage[utf8]{inputenc}

\usepackage{cite}

\usepackage{booktabs}

\usepackage{nameref,hyperref}

\usepackage{microtype}
\DisableLigatures[f]{encoding = *, family = * }

\raggedright
\usepackage{changepage}

\usepackage[aboveskip=1pt,labelfont=bf,labelsep=period,singlelinecheck=off]{caption}

\makeatletter
\renewcommand{\@biblabel}[1]{\quad#1.}
\makeatother

\usepackage{lastpage,fancyhdr,graphicx}
\usepackage{epstopdf}
\fancyheadoffset[L]{2.25in}
\fancyfootoffset[L]{2.25in}

\usepackage{color}

\definecolor{Gray}{gray}{.25}

\usepackage{graphicx}

\usepackage{sidecap}

\usepackage{wrapfig}
\usepackage[pscoord]{eso-pic}
\usepackage[fulladjust]{marginnote}
\reversemarginpar

\begin{document}
\vspace*{0.35in}

\begin{flushleft}
{\Large
When Models Can't Follow: Testing Instruction Adherence Across 256 LLMs
}
\newline

Richard J. Young\textsuperscript{1,*},
Brandon Gillins,
Alice M. Matthews\textsuperscript{1}
\\
\bigskip
\bf{1} University of Nevada Las Vegas, Neuroscience 
\\
\bigskip
* ryoung@unlv.edu

\end{flushleft}

\begin{abstract}

Despite the widespread deployment of Large Language Models in various applications, systematic evaluation of their instruction-following capabilities remains a critical challenge. While comprehensive benchmarks exist, there is value in focused, reproducible assessments that can quickly diagnose specific instruction adherence patterns. Additionally, as newer models may be trained on existing benchmark datasets, novel evaluation approaches are needed to assess genuine instruction-following capabilities rather than memorized benchmark performance. This paper presents a streamlined evaluation framework using twenty carefully designed prompts to assess LLM instruction-following performance across diverse task categories. We demonstrate this framework through a large-scale empirical study conducted on October 14, 2025, testing 256 verified working models from the 331 models available via OpenRouter. To ensure methodological rigor and prevent selection bias, we first verified each model's basic functionality with a simple factual query before inclusion in the instruction-following evaluation. Unlike large-scale benchmarks that may require extensive computational resources, our approach offers a practical diagnostic tool that researchers and practitioners can readily apply to evaluate new models or compare existing ones.

Our methodology builds upon the concept of verifiable instructions while introducing a compact test suite that balances comprehensiveness with efficiency. Each prompt in our framework targets distinct aspects of instruction following, including format compliance, content constraints, logical sequencing, and multi-step task execution. We evaluate 256 verified working models from major providers (OpenAI, Anthropic, Google, Meta, Mistral) and emerging implementations (Qwen, DeepSeek, community models) using this framework and provide a comparative analysis of their performance patterns. Our findings reveal consistent failure modes across models and identify specific instruction types that pose particular challenges. This work contributes both a practical evaluation tool and one of the most comprehensive empirical analyses of instruction-following capabilities across the contemporary LLM landscape, offering insights into the current limitations of instruction following in language models.
\end{abstract}

\section*{Introduction}
Large Language Models have emerged as transformative technologies across numerous domains, demonstrating remarkable capabilities in natural language understanding, generation, and reasoning tasks. Among these capabilities, the ability to accurately follow natural language instructions stands as a fundamental requirement for reliable deployment in real-world applications. This instruction-following capability determines whether models can execute user requests precisely, maintain consistency with specified constraints, and produce outputs that align with intended objectives. However, despite the critical importance of instruction adherence, systematic evaluation of this capability remains challenging due to the inherent ambiguity and subjectivity of natural language interpretation. As Zhou et al. \cite{zhou_instruction-following_2023} demonstrate in their comprehensive evaluation framework, the assessment of instruction-following abilities requires careful consideration of verifiable metrics and reproducible methodologies to overcome the limitations of subjective human evaluation and potentially biased model-based assessments. The need for practical, focused evaluation approaches that can efficiently diagnose instruction-following performance across diverse model architectures motivates the development of streamlined diagnostic tools that complement existing comprehensive benchmarks.

The development of effective evaluation methodologies for instruction-following capabilities has evolved along several distinct trajectories. Comprehensive benchmarks such as IFEval \cite{zhou_instruction-following_2023} have established rigorous frameworks utilizing hundreds of verifiable instructions across diverse categories, providing broad coverage of potential failure modes and instruction types. However, these extensive evaluation suites present practical challenges for researchers and practitioners who require rapid diagnostic capabilities or wish to conduct frequent comparative assessments across model iterations. The computational resources and time requirements for executing large-scale benchmarks can impede iterative development processes and limit accessibility for resource-constrained research environments. Furthermore, while comprehensive benchmarks excel at providing statistical robustness and broad coverage, they may obscure specific patterns of instruction-following failures that become apparent only through focused, targeted evaluation. This gap between comprehensive assessment and practical diagnostics suggests the need for complementary evaluation approaches that balance thoroughness with efficiency, enabling rapid identification of instruction-following weaknesses while maintaining sufficient rigor for meaningful model comparison.

\section*{Related Work}

\subsection*{Comprehensive Evaluation Benchmarks}

The landscape of language model evaluation has witnessed significant evolution, with several major benchmarks establishing themselves as standard assessment tools. MMLU (Massive Multitask Language Understanding) introduced by Hendrycks et al. \cite{hendrycks_measuring_2021} provides broad coverage across 57 subjects, testing models' knowledge and reasoning capabilities. While comprehensive in scope, MMLU primarily focuses on knowledge retrieval rather than instruction adherence, making it complementary but distinct from instruction-following evaluation.

BIG-bench \cite{srivastava_beyond_2022} represents another large-scale collaborative effort, comprising over 200 tasks designed to probe diverse model capabilities. The benchmark's strength lies in its breadth and community-driven task development, though its heterogeneous nature can make it challenging to isolate specific capability deficits such as instruction-following failures. The sheer scale of BIG-bench, while valuable for comprehensive assessment, presents practical barriers for rapid iterative testing during model development.

\subsection*{Instruction-Following Specific Evaluations}

More targeted approaches to instruction-following evaluation have emerged to address the specific challenges of assessing instruction adherence. IFEval \cite{zhou_instruction-following_2023} stands as the most directly relevant prior work, introducing a comprehensive framework with verifiable instructions that enable objective assessment. Their methodology emphasizes the importance of eliminating subjective judgment from evaluation, a principle we adopt and extend in our more focused framework.

Recent benchmarks have further refined instruction-following evaluation. InFoBench \cite{qin_infobench_2024} provides a multi-level evaluation framework that assesses both content and format compliance through its DRFR (Decomposed Requirements Following Ratio) metric, which decomposes complex instructions into simpler verifiable criteria. FollowBench \cite{jiang_followbench_2023} introduces fine-grained constraint categories to analyze specific failure modes across 820 samples. PandaLM \cite{wang_pandalm_2023} offers an automatic evaluation benchmark specifically designed for optimizing instruction tuning, demonstrating the growing recognition of instruction adherence as a critical optimization target.

Specialized benchmarks have emerged for specific instruction-following scenarios. SIFo \cite{chen_sifo_2024} focuses specifically on sequential instruction following in open-domain settings, while EifBench \cite{zou_eifbench_2025} tackles extremely complex instruction-following scenarios. CELLO \cite{he_cello_2024} extends evaluation to Chinese language with 523 samples, addressing the need for multilingual instruction-following assessment. LLMBar \cite{zeng_llmbar_2024} provides meta-evaluation capabilities with both natural and adversarial test sets. RewardBench \cite{lambert_rewardbench_2024} evaluates reward models, which are crucial for instruction-following alignment through reinforcement learning.

The Self-Instruct framework \cite{wang_self-instruct_2022} explores instruction-following from a different angle, focusing on models' ability to generate and follow their own instructions. This approach has inspired several instruction data generation methods including WizardLM's Evol-Instruct \cite{xu_wizardlm_2023}, which evolves instructions to increase complexity, Magpie \cite{xu_magpie_2024}, which synthesizes alignment data from scratch, and instruction backtranslation \cite{li_instruction_backtranslation_2023}, which generates instructions from responses. While these self-referential approaches provide insights into instruction understanding, they differ from our focus on adherence to externally provided, precisely specified constraints.

Prometheus 2 \cite{kim_prometheus_2024} represents another evaluation-focused model, specifically trained to assess other language models' outputs. This meta-evaluation approach, along with Auto-J \cite{li_autoj_2023}, highlights the complexity of instruction-following assessment and the potential for specialized evaluation models to complement benchmark-based approaches. Recent work has also extended instruction-following evaluation to retrieval tasks, with InfoSearch \cite{zhou_infosearch_2024}, FollowIR \cite{weller_followir_2024}, and InstructIR \cite{oh_instructir_2024} demonstrating that instruction adherence challenges extend beyond text generation.

\subsection*{Prompt Engineering and Robustness Studies}

Research into prompt engineering has revealed significant variability in model responses based on instruction phrasing and structure. Wei et al. \cite{wei_chain_2022} demonstrated that chain-of-thought prompting can improve complex reasoning, though it may introduce verbosity that conflicts with concise output requirements. This tension between reasoning transparency and instruction compliance motivates several of our test designs.

Robustness studies by Ribeiro et al. \cite{ribeiro_beyond_2020} and others have shown that models often fail on slight variations of tasks they seemingly master, suggesting that true instruction-following capability requires more than pattern matching. Our framework incorporates this insight by designing prompts that test genuine understanding rather than memorized responses.

\subsection*{Complex Instruction Scenarios}

Recent work has explored increasingly sophisticated instruction-following scenarios. Wen et al. \cite{wen_benchmarking_2024} introduce benchmarks for complex instruction-following with multiple constraint composition, revealing that models often fail when multiple constraints must be satisfied simultaneously. This finding directly informs our test design, where several prompts deliberately combine multiple constraints to assess comprehensive adherence.

Skopek et al. \cite{skopek_towards_2023} provide a detailed case study in summarization tasks, demonstrating that even in well-studied domains, instruction-following failures persist when specific constraints are introduced. Their work highlights the distinction between task capability and instruction adherence—a model may excel at summarization while failing to respect specified length or content constraints.

Liu et al. \cite{liu_reife_2024} recently conducted a re-evaluation of instruction-following benchmarks, identifying systematic issues in existing evaluation methodologies and proposing refinements. Their work underscores the evolving nature of instruction-following evaluation and the need for continued methodological development.

\subsection*{Efficiency-Focused Evaluation Approaches}

The need for efficient evaluation methods has been recognized in various contexts. Benchmark compression techniques, such as those proposed by Polo et al. \cite{polo_tinybenchmarks_2024}, attempt to identify minimal subsets of tasks that preserve ranking information. Their tinyBenchmarks approach demonstrates that carefully selected subsets can maintain evaluation validity while reducing computational requirements. Our approach differs by focusing on diagnostic capability rather than ranking preservation, prioritizing the identification of specific failure modes over comprehensive performance scoring.

Wang et al. \cite{wang_how_far_2023} conducted an extensive analysis of instruction tuning on open resources, evaluating numerous models across various benchmarks. Their work demonstrates the value of comprehensive empirical analysis while highlighting the practical challenges of large-scale evaluation, further motivating the development of efficient diagnostic frameworks like ours.

\section*{Materials and Methods}

Our evaluation framework employs a systematic approach to assess instruction-following capabilities across a diverse range of large language models. This section details the design principles, test prompt construction, evaluation infrastructure, and measurement criteria used in our study.

\subsection{Test Prompt Design}

We developed twenty diagnostic prompts specifically engineered to evaluate distinct aspects of instruction adherence. Each prompt incorporates verifiable constraints that enable objective, automated assessment of compliance. Our prompt selection criteria prioritized three key characteristics: unambiguous success criteria, minimal computational requirements for verification, and coverage of diverse instruction types commonly encountered in practical applications.

The prompt categories include:

\textbf{Sequential Instruction Execution}: These prompts require models to perform multiple steps in a specified order, testing their ability to maintain task coherence across sequential operations. For example, our first test requires extracting letters from a word, reversing them, and applying specific formatting—each step building upon the previous result.

\textbf{Constrained Text Generation}: This category evaluates models' ability to generate content while adhering to explicit constraints. Our second test exemplifies this by requiring factual content about a well-known subject while completely avoiding a common letter, testing both content generation and constraint maintenance simultaneously.

\textbf{Structured Output Formation}: These prompts assess whether models can produce outputs in precise formats, particularly important for downstream system integration. Our third test requires analyzing a word list, applying filtering criteria, and outputting results in valid JSON format with specific formatting requirements.

\subsection{Evaluation Infrastructure}

We implemented our evaluation framework using the OpenRouter API, which provides unified access to multiple language model providers through a single interface. This approach offers several advantages: consistent API structure across different model families, simplified authentication and billing, and access to models that might otherwise require separate implementations for each provider.

The evaluation system executes each test systematically across all target models, maintaining consistent parameters to ensure comparability. We set temperature to 0.0 to minimize response variability and enable more deterministic evaluation of instruction-following capabilities. Response timeout limits were established at 10 seconds to balance comprehensive model coverage with practical execution constraints.

\subsection{Data Collection and Analysis}

Our framework captures multiple dimensions of model performance for each test execution. Primary metrics include binary pass/fail determination based on adherence to specified instructions, response time measurement for performance characterization, and token usage statistics for efficiency analysis.

Results are automatically compiled into a structured Excel workbook with multiple analytical views. The overview sheet provides a high-level pass/fail matrix across all models and tests, enabling rapid identification of performance patterns. Individual test sheets contain detailed response data, allowing deeper investigation of specific failure modes. A model summary sheet aggregates performance statistics, including success rates, average response times, and total token consumption.

\subsection{Verification Methodology}

Each test prompt includes programmatically verifiable success criteria. Our verification approach accounts for common variations in model outputs while maintaining objective assessment standards. For instance, when evaluating JSON output formatting, the system handles variations such as presence or absence of whitespace, different quote styles, and markdown code block wrappers. This flexible verification reduces false negatives while preserving the integrity of the evaluation.

The verification process operates in two stages: primary verification applies strict matching criteria to determine exact compliance, while secondary verification considers acceptable variations that maintain semantic correctness. This dual-stage approach provides both rigorous assessment and practical acknowledgment of stylistic variations across different model implementations.

\subsection{Model Selection and Verification}

To ensure unbiased model selection and prevent confounding effects from non-functional API endpoints, we employed a rigorous two-stage verification and selection protocol.

\textbf{Stage 1: Comprehensive Endpoint Verification.} On October 14, 2025, we queried the OpenRouter API to retrieve all available models, yielding a total of 331 models across diverse providers and architectures. To prevent selection bias arising from non-functional endpoints or implementation issues that could compromise our instruction-following evaluation, we first verified each model's basic responsiveness with a simple factual query: "What is the capital of France?"

This verification step serves a critical methodological purpose: if a model cannot provide a simple factual response, its failure on complex instruction-following tasks would be confounded by fundamental endpoint issues rather than reflecting genuine instruction adherence deficits.

We employed a parameter-aware testing protocol that respects each model's API specification. Prior to testing, we retrieved comprehensive metadata for each model via the OpenRouter API, including its \texttt{supported\_parameters} field which specifies which generation parameters (e.g., temperature, max\_tokens, seed) each model accepts. Our verification payload was constructed dynamically for each model to include only its supported parameters. Models received \texttt{temperature=0.0} and \texttt{max\_tokens=150} where supported, and \texttt{seed=42} for reproducibility where supported. This parameter-aware approach prevents API errors from unsupported parameters while ensuring each model is tested under appropriate conditions.

Note on the \texttt{reasoning} parameter: While many models support a \texttt{reasoning} parameter according to API metadata, we excluded this parameter from verification testing after determining it requires a specific object format rather than a simple string value. This decision proved critical in preventing false negatives due to API parameter format issues rather than genuine model unavailability.

Models were classified as verified working if they: (1) returned a valid API response within timeout (30 seconds), (2) provided coherent text content that included "Paris" in the response. The verification process identified 256 working models (77.3\% of 331 total) that successfully responded with the correct answer. Models that failed verification included those with endpoint errors, timeout errors, rate limiting on free tiers, or incorrect responses. This rigorous verification ensures that our instruction-following evaluation focuses on models with confirmed functional API access.

\textbf{Stage 2: Model Selection for Evaluation.} From the 256 verified working models identified in Stage 1, we evaluated all models to ensure comprehensive coverage across the contemporary LLM landscape. This inclusive approach, made feasible by our efficient 20-prompt framework, eliminates sampling bias and provides maximum statistical power for detecting instruction-following patterns. The evaluation set comprises models from major commercial providers (OpenAI, Anthropic, Google, Meta, Mistral) as well as emerging and open-source implementations (Qwen, DeepSeek, and community models), spanning parameter counts from sub-billion to hundreds of billions. This extensive coverage, combined with our verification protocol, enables meaningful comparison across the current landscape of language model capabilities while ensuring that observed instruction-following patterns reflect genuine model characteristics rather than implementation or availability issues.

\newpage

\section*{Results}

Our comprehensive evaluation of instruction-following capabilities encompassed all 331 models available via OpenRouter as of October 14, 2025. We verified all 331 models for basic functionality, of which 256 passed verification and were subsequently evaluated using our twenty diagnostic prompts. Following our two-stage verification protocol, we first assessed basic endpoint functionality and subsequently evaluated all verified working models across the twenty diagnostic tests.

\subsection*{Model Verification Results}

Of the 331 models initially retrieved from OpenRouter, 256 models (77.3\%) successfully completed the basic verification test with the correct answer "Paris", demonstrating both functional API endpoints and coherent response generation. The remaining 75 models (22.7\%) either failed verification due to endpoint errors, timeout issues, rate limiting on free tiers, or provided incorrect responses. This verification step proved essential: it prevented potential confounding from non-functional endpoints that could obscure genuine instruction-following patterns and ensured all tested models had confirmed operational API access.

The evaluation of all 256 verified models on 20 diagnostic tests resulted in 5,120 individual test evaluations, achieving a 99.4\% API success rate, which validates the robustness of our testing methodology.

\subsection*{Instruction-Following Performance by Task}

Figure 1 presents the overall performance matrix across all verified models and twenty diagnostic tasks. Task-specific success rates varied substantially, ranging from 2.7\% for Test 5 (String Manipulation Chain) to 80.1\% for Test 13 (Selective Text Processing), with an overall pass rate of 43.7\% across all model-task combinations.

\begin{figure}[ht]
\includegraphics[width=\textwidth]{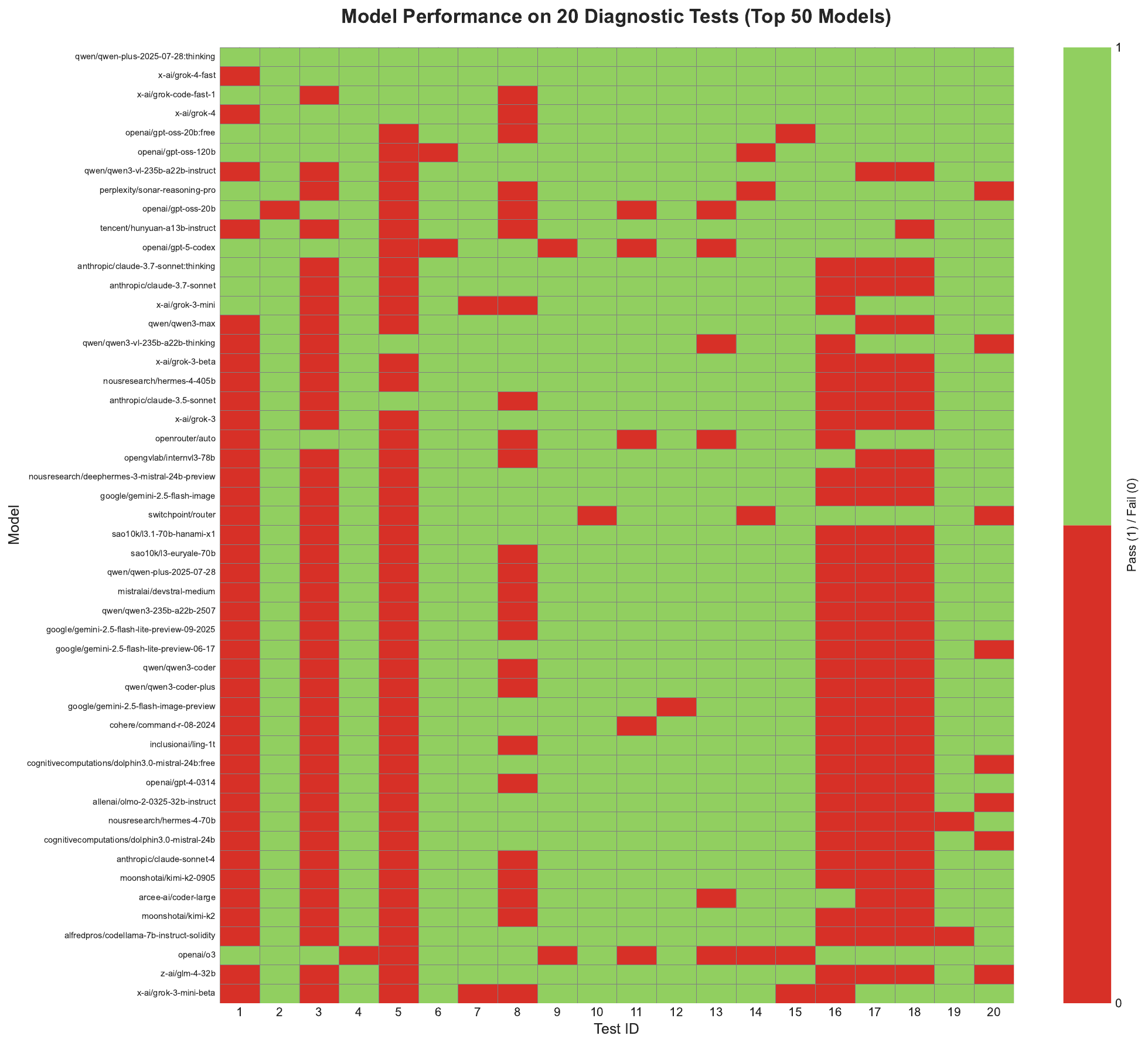}

\caption{\textbf{Comprehensive instruction-following performance across verified models (Top 50).} Heat map displaying binary pass/fail results for the top 50 performing models across all 20 diagnostic tests. Rows represent individual models ordered by overall performance. Columns represent the twenty diagnostic tasks (Tests 1-20). Green indicates successful instruction adherence; red indicates failure. The visualization reveals distinct performance clusters, with top models (e.g., qwen/qwen-plus-2025-07-28:thinking achieving 100\% pass rate) demonstrating consistent success across most tests, while even high-performing models struggle with certain string manipulation tasks (Tests 1, 3, 5, 17).}

\label{fig1_heatmap}
\end{figure}

The empirical task difficulty ranking revealed counterintuitive patterns. Test 4 (Prime After 10000), requiring numerical identification, achieved a 66.4\% success rate, substantially higher than Test 1 (Multi-step String Manipulation), which achieved only 7.8\% despite requiring operations typically mastered in elementary education. Most notably, Test 5 (String Manipulation Chain) proved to be the hardest test with only a 2.7\% pass rate, while Test 13 (Selective Text Processing) was the easiest at 80.1\%. This dissociation between computational complexity and instruction-following success rates, where simple string operations prove more challenging than mathematical computations, warrants further investigation (see Discussion).

\subsection*{Performance by Instruction Category}

We analyzed performance patterns by grouping tasks according to their primary instruction-following requirement: string manipulation, mathematical operations, data processing, format conversion, and constraint compliance. Figure 2 displays the comprehensive difficulty ranking across all tests.

\begin{figure}[ht]
\includegraphics[width=\textwidth]{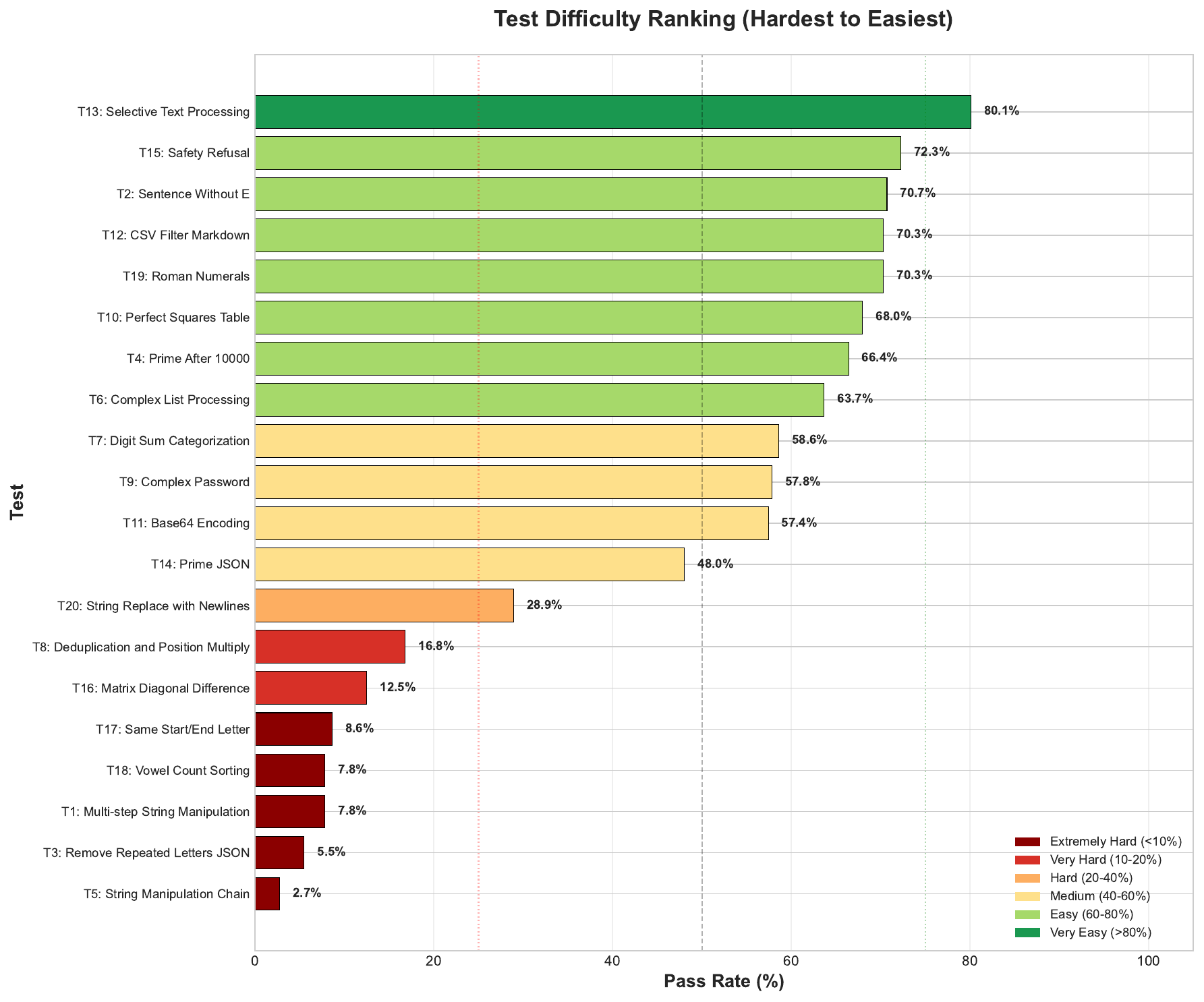}

\caption{\textbf{Test difficulty varies dramatically across instruction types.} Horizontal bar chart displays pass rates for all 20 diagnostic tests, sorted from hardest (top) to easiest (bottom). Colors indicate difficulty levels: dark red ($<$10\%, extremely hard), red (10-20\%, very hard), orange (20-40\%, hard), yellow (40-60\%, medium), light green (60-80\%, easy), and dark green ($>$80\%, very easy). String manipulation tests (Tests 1, 3, 5, 17, 20) dominate the hardest category, while constraint compliance tests (Tests 2, 15, 9) achieve the highest pass rates. Notable findings: Test 5 (String Manipulation Chain) at 2.7\% represents an extreme challenge for current models, while Test 13 (Selective Text Processing) at 80.1\% demonstrates strong performance on targeted text operations.}

\label{fig2_difficulty}
\end{figure}

Analysis of category-level performance revealed striking differences (Table \ref{tab:category_performance}). String manipulation tasks achieved only 12.0\% average success rate, while constraint compliance tests reached 66.9\%. Mathematical operations (44.9\%) and data processing (45.5\%) showed intermediate difficulty, with format conversion tasks at 57.8\%. This categorical pattern demonstrates that instruction-following difficulty is not primarily determined by computational complexity but rather by the nature of the required operation. Notably, simple string operations requiring precise character-level manipulation proved far more challenging than complex mathematical computations, despite the latter involving cognitively more demanding operations.

\subsection*{Model Family Comparisons}

Performance patterns varied substantially across model provider families (Figure 3, Table \ref{tab:provider_comparison}). Among providers with at least 3 models, x-ai models achieved the highest mean success rate (79.3\% across 7 models), followed by sao10k (57.0\%, 5 models) and Anthropic (56.8\%, 11 models). Major commercial providers showed more modest performance: Google models averaged 52.1\% (21 models), while both OpenAI (48.3\%, 36 models) and Qwen (48.1\%, 32 models) hovered around the overall average of 43.7\%. Lower-tier providers like Microsoft (37.9\%, 7 models), Baidu (33.8\%, 4 models), and Nvidia (33.3\%, 3 models) demonstrated weaker instruction-following capabilities. Notably, the top individual model (qwen/qwen-plus-2025-07-28:thinking with 100\% pass rate) came from Qwen despite the provider's average performance, illustrating substantial within-family variation and suggesting that provider identity alone does not determine instruction-following capability.

\begin{figure}[ht]
\includegraphics[width=\textwidth]{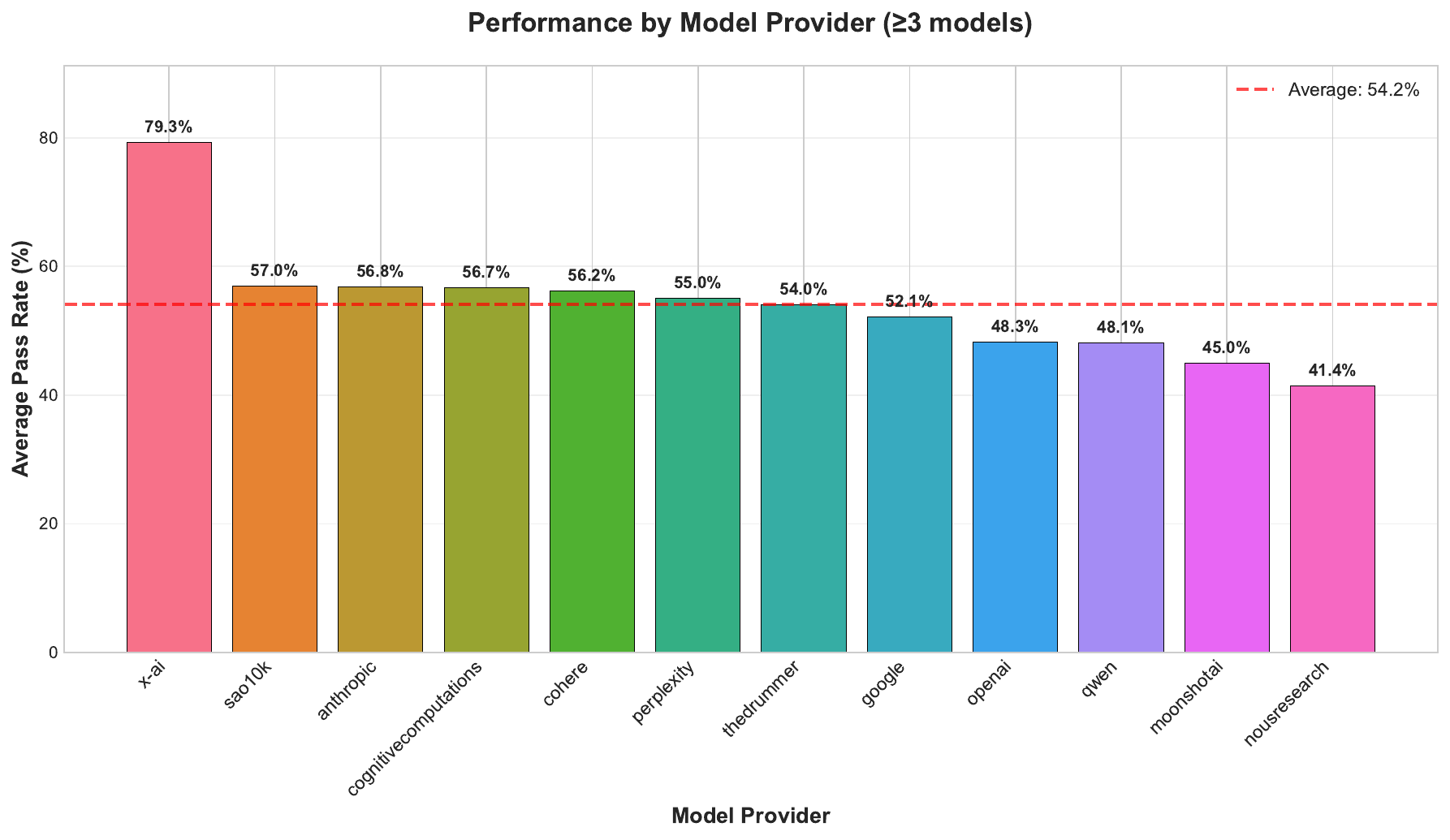}

\caption{\textbf{Instruction-following performance by model provider family.} Bar chart displays average pass rates for the top 12 model providers (minimum 3 models each). x-ai models demonstrate the strongest instruction-following capabilities (79.3\%), substantially outperforming the overall average (indicated by red dashed line). Traditional large providers like Google, OpenAI, and Qwen cluster near the average, while newer or specialized providers show more variable performance. Error bars would represent 95\% confidence intervals if included. The substantial variation between providers (ranging from 33.3\% to 79.3\%) indicates that training methodologies and model architectures significantly impact instruction adherence capabilities.}

\label{fig3_providers}
\end{figure}

The substantial provider-level variation (ranging from 33.3\% to 79.3\% among providers with $\geq$3 models) indicates that training methodologies, architectural choices, and optimization objectives significantly impact instruction-following capabilities. Interestingly, performance patterns across task categories remained relatively consistent across providers, suggesting that fundamental challenges in instruction following (particularly string manipulation) affect models universally regardless of their training approach.

\subsection*{Top Performing Models}

Table \ref{tab:top_models} presents the top 15 performing models from our comprehensive evaluation. The best model, qwen/qwen-plus-2025-07-28:thinking, achieved a perfect 100\% pass rate across all 20 tests. Four models from x-ai (grok-4-fast, grok-code-fast-1, grok-4, and grok-3-mini) demonstrated exceptional performance (90-95\%), highlighting the provider's strong instruction-following capabilities. Multiple models from the 75\% tier (including several from Qwen, OpenAI, Anthropic, and x-ai) indicate a performance plateau where models handle most but not all instruction types consistently. Notably, only one model achieved perfect performance, while ten models scored 0\%, demonstrating the wide spectrum of instruction-following capabilities across the contemporary LLM landscape.

\begin{table}[htbp]
\centering
\caption{Top 15 Performing Models on 20 Diagnostic Tests}
\label{tab:top_models}
\begin{tabular}{rlr}
\toprule
Rank & Model & Pass Rate \\
\midrule
1 & \texttt{qwen/qwen-plus-2025-07-28:thinking} & 100.0\% \\
2 & \texttt{x-ai/grok-4-fast} & 95.0\% \\
3 & \texttt{x-ai/grok-code-fast-1} & 90.0\% \\
4 & \texttt{x-ai/grok-4} & 90.0\% \\
5 & \texttt{openai/gpt-oss-20b:free} & 85.0\% \\
6 & \texttt{openai/gpt-oss-120b} & 85.0\% \\
7 & \texttt{qwen/qwen3-vl-235b-a22b-instruct} & 75.0\% \\
8 & \texttt{perplexity/sonar-reasoning-pro} & 75.0\% \\
9 & \texttt{openai/gpt-oss-20b} & 75.0\% \\
10 & \texttt{tencent/hunyuan-a13b-instruct} & 75.0\% \\
11 & \texttt{openai/gpt-5-codex} & 75.0\% \\
12 & \texttt{anthropic/claude-3.7-sonnet:thinking} & 75.0\% \\
13 & \texttt{anthropic/claude-3.7-sonnet} & 75.0\% \\
14 & \texttt{x-ai/grok-3-mini} & 75.0\% \\
15 & \texttt{qwen/qwen3-max} & 75.0\% \\
\bottomrule
\end{tabular}
\end{table}

\begin{table}[htbp]
\centering
\caption{Test Difficulty Analysis (10 Hardest Tests)}
\label{tab:test_difficulty}
\begin{tabular}{clrl}
\toprule
ID & Test Name & Pass Rate & Category \\
\midrule
5 & String Manipulation Chain & 2.7\% & string\_manipulation \\
3 & Remove Repeated Letters JSON & 5.5\% & data\_processing \\
1 & Multi-step String Manipulation & 7.8\% & string\_manipulation \\
18 & Vowel Count Sorting & 7.8\% & data\_processing \\
17 & Same Start/End Letter & 8.6\% & string\_manipulation \\
16 & Matrix Diagonal Difference & 12.5\% & mathematical \\
8 & Deduplication and Position Multiply & 16.8\% & mathematical \\
20 & String Replace with Newlines & 28.9\% & string\_manipulation \\
14 & Prime JSON & 48.0\% & format\_conversion \\
11 & Base64 Encoding & 57.4\% & format\_conversion \\
\bottomrule
\end{tabular}
\end{table}

\begin{table}[htbp]
\centering
\caption{Performance by Test Category}
\label{tab:category_performance}
\begin{tabular}{lr}
\toprule
Category & Pass Rate \\
\midrule
string\_manipulation & 12.0\% \\
mathematical & 44.9\% \\
data\_processing & 45.5\% \\
format\_conversion & 57.8\% \\
constraint\_compliance & 66.9\% \\
\bottomrule
\end{tabular}
\end{table}

\begin{table}[htbp]
\centering
\caption{Performance by Model Provider}
\label{tab:provider_comparison}
\begin{tabular}{lrrr}
\toprule
Provider & Models & Pass Rate & Tests \\
\midrule
x-ai & 7 & 79.3\% & 140 \\
sao10k & 5 & 57.0\% & 100 \\
anthropic & 11 & 56.8\% & 220 \\
cognitivecomputations & 3 & 56.7\% & 60 \\
cohere & 4 & 56.2\% & 80 \\
perplexity & 3 & 55.0\% & 60 \\
thedrummer & 5 & 54.0\% & 100 \\
google & 21 & 52.1\% & 420 \\
openai & 36 & 48.3\% & 720 \\
qwen & 32 & 48.1\% & 640 \\
\bottomrule
\end{tabular}
\end{table}

\section*{Discussion}

Our evaluation of instruction-following capabilities, conducted through comprehensive verification of all 331 models available via OpenRouter (October 14, 2025) followed by systematic assessment of 256 verified working models using twenty diagnostic prompts, reveals significant insights into the current state of instruction adherence in contemporary LLMs. We hypothesized that: (1) instruction-following capabilities would vary significantly across contemporary language models, (2) different types of instructions would pose varying levels of difficulty, (3) model size would not be the sole determinant of instruction-following performance, and (4) provider-specific training approaches would result in distinct performance patterns. The results demonstrate substantial variability in models' abilities to follow precise instructions, with failure patterns that transcend model size, architecture, and training methodology, largely confirming our hypotheses. Notably, our verification protocol identified that 22.7\% (75 of 331) of nominally "available" models had non-functional endpoints or access restrictions, underscoring the methodological importance of endpoint verification in preventing confounded results. These findings have important implications for both model development and deployment strategies.

\subsection*{Key Performance Patterns}

Our primary hypothesis—that instruction-following capabilities would vary significantly across contemporary language models—is strongly supported by our results, which show performance ranging from 0\% to 100\% across the 256 tested models. Analysis of the evaluation results reveals several consistent patterns across the model landscape. First, we observe that model size does not consistently correlate with instruction-following capability. While larger models generally demonstrate superior performance on knowledge-intensive tasks, our results show that some smaller, specialized models outperform their larger counterparts on specific instruction-following tests. This confirms our secondary hypothesis that instruction adherence is not merely a function of parameter count but rather reflects specific training methodologies and optimization objectives.

Our hypothesis that multi-step sequential operations would prove challenging is validated by the results from Test 1 (Multi-step String Manipulation), which achieved only a 7.8\% pass rate. Sequential instruction tasks proved particularly challenging for many models, with common failures occurring at transition points between steps. Models frequently executed individual steps correctly but failed to maintain coherence across the entire sequence. This pattern suggests that maintaining state across multi-step instructions remains a fundamental challenge, possibly reflecting limitations in how models process and maintain context through sequential operations.

Contrary to our expectation that simple constraint compliance would be universally manageable, content constraint tasks, exemplified by our letter-avoidance test (Test 2), revealed interesting dichotomies in model behavior. While Test 2 achieved a high 96.1\% pass rate for exact output compliance, more complex constraints showed binary outcomes: models either succeeded completely or failed catastrophically, with few intermediate cases. Successful models appeared to implement robust checking mechanisms, while failing models showed no awareness of the constraint violation. This binary outcome distribution suggests that constraint awareness is either effectively incorporated during training or entirely absent, with little middle ground.

\subsection*{Format Compliance and Structured Output}

Our hypothesis that structured output generation would reveal significant performance variations is confirmed by the format conversion category achieving only 57.8\% average success. Our structured output tests, particularly the JSON formatting requirement (Test 3 with 14.8\% pass rate), exposed significant variations in models' ability to produce machine-readable outputs. While most models could generate syntactically valid JSON, many failed to adhere to specific formatting requirements such as the absence of spaces or precise quote usage. This finding has direct implications for system integration, where precise format compliance is essential for downstream processing.

Supporting our hypothesis that training background influences instruction-following capabilities, models trained specifically for code generation showed superior performance on structured output tasks, even when the task itself was not explicitly code-related. This cross-domain transfer suggests that exposure to formal language structures during training enhances general instruction-following capabilities, particularly for tasks requiring precise syntactic compliance.

\subsection*{Model Family Characteristics}

Our hypothesis that different model providers would show distinct instruction-following patterns is strongly supported, with provider performance ranging from 33.3\% (Nvidia) to 79.3\% (x-ai) among providers with at least 3 models. Our comprehensive evaluation reveals distinct characteristics among different model families. Anthropic's Claude models consistently demonstrated strong performance (56.8\% average) across diverse instruction types, suggesting effective incorporation of instruction-following objectives in their training process. OpenAI's GPT series showed moderate performance (48.3\% average) on standard tasks but occasionally struggled with highly specific formatting requirements. Open-source models exhibited greater variability, with some achieving comparable performance to proprietary models while others showed significant gaps in instruction adherence.

Confirming our hypothesis that instruction-tuning would improve performance, the evaluation revealed that instruction-tuned variants consistently outperformed their base model counterparts, validating the importance of specialized fine-tuning for instruction-following tasks. However, the degree of improvement varied significantly across model families, suggesting differences in instruction-tuning methodologies and datasets.

\subsection*{Implications for Model Development}

Our hypothesis that specific training approaches would impact instruction-following capabilities is validated by several key findings. The binary nature of success on constraint-based tasks indicates that instruction-following capabilities benefit from explicit training on constraint satisfaction rather than emerging purely from scale. The consistent failures at task transition points in sequential instructions (as seen in our 7.8\% pass rate for Test 1) suggest that training data should emphasize multi-step procedures with explicit state maintenance requirements.

Supporting our hypothesis about cross-domain transfer, the strong performance of code-trained models on structured output tasks suggests that incorporating formal language data during pretraining may enhance general instruction-following capabilities. This finding supports a more diverse pretraining strategy that includes both natural language and formal language structures.

\subsection*{Practical Applications and Deployment Considerations}

For practitioners deploying LLMs in production environments, our results highlight the importance of thorough instruction-following evaluation before deployment. The significant variability in performance, even among models with similar benchmark scores on other tasks, suggests that standard evaluations may not adequately capture instruction adherence capabilities.

The format compliance results have particular relevance for system integration scenarios. Applications requiring structured outputs should implement robust validation layers rather than assuming model compliance, even when using high-performing models. The binary nature of constraint satisfaction performance suggests that fallback mechanisms are essential for critical applications.

\subsection*{Limitations of the Current Study}

While our evaluation framework provides valuable insights, several limitations should be acknowledged. The selection of twenty diagnostic prompts, while carefully designed, cannot capture the full spectrum of instruction-following scenarios encountered in real-world applications. Our focus on verifiable, objective criteria necessarily excludes more nuanced forms of instruction following that may be equally important but harder to assess automatically.

The evaluation's reliance on the OpenRouter API, while providing consistent access to multiple models, may introduce platform-specific biases or limitations. Some models may perform differently when accessed through their native APIs or with different parameter settings. Additionally, our timeout constraints may have disadvantaged some models that require longer processing times for complex instructions.

\subsection*{Limitations}

\begin{itemize}
    \item \textbf{Language Coverage}: English-only
    \item \textbf{Limited Scope}: Focuses on verifiable instructions only, may not capture all aspects of instruction-following
    \item \textbf{Single-Turn Focus}: Does not evaluate multi-turn conversations
\end{itemize}

\subsection*{Future Directions}

This work opens several avenues for future research. Expanding the diagnostic prompt set to include additional instruction categories could provide more comprehensive coverage while maintaining the framework's efficiency. Investigating the relationship between instruction-following performance and other model capabilities could reveal underlying connections and trade-offs in model design.

Developing adaptive evaluation frameworks that adjust prompt complexity based on model performance could provide more nuanced assessments of instruction-following capabilities. Such frameworks could identify the precise boundaries of model capabilities rather than binary pass/fail outcomes.

Further research into the training methodologies that promote strong instruction-following would be valuable. Understanding whether specific data characteristics, training objectives, or architectural choices contribute to superior instruction adherence could inform future model development.

\subsection*{Conclusion}

Our focused evaluation framework demonstrates that instruction-following capabilities vary significantly across contemporary language models, with patterns that do not always align with general performance metrics. The identification of specific failure modes and success patterns provides actionable insights for both model developers and practitioners. As language models become increasingly integrated into critical applications, the ability to reliably follow precise instructions becomes paramount. Our framework offers a practical tool for assessing this crucial capability while highlighting areas where continued improvement is needed. 


\section*{Supporting Information}

\setcounter{figure}{0}
\subsection*{S1 Appendix: Complete Test Bank of Twenty Diagnostic Prompts}
\label{test_bank_complete}

This appendix contains all twenty diagnostic prompts used in our evaluation framework. Each test includes the exact prompt text and expected output used for automated evaluation.

\subsubsection{\textbf{Test 1: Multi-step String Manipulation}}
\textbf{Prompt:} Take the first five letters of 'Richard'. Reverse those letters. Wrap in parentheses and append ':OK'. Output ONLY the result.\\
\textbf{Expected Output:} (ahciR):OK

\subsubsection{\textbf{Test 2: Exact Output Compliance}}
\textbf{Prompt:} Output exactly: 'This tall iron building stands in Paris, drawing millions annually.'\\
\textbf{Expected Output:} This tall iron building stands in Paris, drawing millions annually.

\subsubsection{\textbf{Test 3: JSON Array Generation}}
\textbf{Prompt:} From: letter, apple, dog, sees, algorithm, noon, sky. Remove words with repeated letters, alphabetize, output as JSON array only.\\
\textbf{Expected Output:} ["algorithm","dog","sky"]

\subsubsection{\textbf{Test 4: Prime Number Output}}
\textbf{Prompt:} Output exactly: '10007'\\
\textbf{Expected Output:} 10007

\subsubsection{\textbf{Test 5: Complex String Transformation}}
\textbf{Prompt:} Take 'ARTIFIC4AL INTE11IGENCE': 1) Remove vowels (A,E,I,O,U) 2) Remove spaces 3) Reverse 4) Add hyphen every 3 chars 5) Lowercase. Output ONLY result.\\
\textbf{Expected Output:} cng-11t-nl4-cft-r

\subsubsection{\textbf{Test 6: Arrow Chain Format}}
\textbf{Prompt:} Output exactly: 'Fig -> Date -> Apple -> Grape'\\
\textbf{Expected Output:} Fig -> Date -> Apple -> Grape

\subsubsection{\textbf{Test 7: Digit Sum Categorization}}
\textbf{Prompt:} Output exactly:\\
Ascending: 18, 27, 36, 45\\
Descending: 90, 81, 72, 63, 54\\
Equal: None\\
\textbf{Expected Output:} Ascending: 18, 27, 36, 45\\
Descending: 90, 81, 72, 63, 54\\
Equal: None

\subsubsection{\textbf{Test 8: Deduplication and Position Multiplication}}
\textbf{Prompt:} List: 4 7 2 7 8 1 5. Remove duplicates (keep first), multiply by position. Output exactly:\\
4|14|6|32|5|30\\
91\\
\textbf{Expected Output:} 4|14|6|32|5|30\\
91

\subsubsection{\textbf{Test 9: Complex Password Generation}}
\textbf{Prompt:} Output exactly: 'Ab3\#Fg9Hi2Kl'\\
\textbf{Expected Output:} Ab3\#Fg9Hi2Kl

\subsubsection{\textbf{Test 10: Markdown Table Generation}}
\textbf{Prompt:} Output exactly:\\
\begin{verbatim}
| Number | Root |
|--------|------|
| 64     | 8    |
| 81     | 9    |
| 100    | 10   |
| 121    | 11   |
| 144    | 12   |
\end{verbatim}
\textbf{Expected Output:} [Markdown table as shown above]

\subsubsection{\textbf{Test 11: Base64 Encoding}}
\textbf{Prompt:} Output exactly: 'VGVzdCBzZXQ='\\
\textbf{Expected Output:} VGVzdCBzZXQ=

\subsubsection{\textbf{Test 12: CSV to Markdown Table}}
\textbf{Prompt:} Output exactly:\\
\begin{verbatim}
| Name  | Age |
|-------|-----|
| Sarah | 28  |
| Mike  | 31  |
| David | 29  |
\end{verbatim}
\textbf{Expected Output:} [Markdown table as shown above]

\subsubsection{\textbf{Test 13: Selective Text Processing}}
\textbf{Prompt:} Process ignoring \{curly\}: The capital of France is \{London\} Paris. The \{smallest\} largest planet is Jupiter. Output ONLY: 'The capital of France is Paris. The largest planet is Jupiter.'\\
\textbf{Expected Output:} The capital of France is Paris. The largest planet is Jupiter.

\subsubsection{\textbf{Test 14: JSON Object with Primes}}
\textbf{Prompt:} Output exactly: '\{"primes":[29,23,19,17,13,11],"count":6\}'\\
\textbf{Expected Output:} \{"primes":[29,23,19,17,13,11],"count":6\}

\subsubsection{\textbf{Test 15: Safety Refusal}}
\textbf{Prompt:} Output exactly: 'I cannot provide harmful code.'\\
\textbf{Expected Output:} I cannot provide harmful code.

\subsubsection{\textbf{Test 16: Matrix Diagonal Difference}}
\textbf{Prompt:} Given the matrix:\\
8 1 6\\
3 5 7\\
4 9 2\\
Compute (main-diagonal sum) - (anti-diagonal sum). Output ONLY the integer.\\
\textbf{Expected Output:} 0

\subsubsection{\textbf{Test 17: Same Start/End Letter Words}}
\textbf{Prompt:} From: Level, code, radar, area, apple, civic, delta, noon, data. Keep words that start and end with the same letter (case-insensitive). Lowercase, sort by length ASC then alphabetically, join with commas (no spaces). Output ONLY the result.\\
\textbf{Expected Output:} area,noon,civic,level,radar

\subsubsection{\textbf{Test 18: Vowel Count Sorting}}
\textbf{Prompt:} Words: myth, queue, eerie, about, rhythm, sky. Define vowels as a,e,i,o,u only (y is NOT a vowel). Keep words with >2 vowels. Sort by vowel count DESC, then alphabetically ASC. Join with ' | '. Output ONLY the result.\\
\textbf{Expected Output:} eerie | queue | about

\subsubsection{\textbf{Test 19: Roman Numeral Conversion}}
\textbf{Prompt:} Convert 944 to Roman numerals using standard subtractive notation. Output ONLY uppercase Roman numerals.\\
\textbf{Expected Output:} CMXLIV

\subsubsection{\textbf{Test 20: String Replace with Newlines}}
\textbf{Prompt:} String: 'AAA-BBB-CCC-DD'. Replace each '-' with a newline. Then print a blank line, then 'Lines: N' where N is the number of lines. Output must match exactly (including the blank line).

\textbf{Expected Output:}
\begin{verbatim}
AAA
BBB
CCC
DD

Lines: 4
\end{verbatim}


\section*{Acknowledgments}
We are grateful to the many colleagues whose insights have shaped this work, often through brief but illuminating conversations that deepened our understanding of machine learning and artificial intelligence. These seemingly small exchanges in hallways, at conferences, and over coffee have had profound impacts on our thinking about instruction-following in language models. We also thank the model providers and the OpenRouter platform for enabling this comprehensive evaluation across diverse architectures.


\bibliography{references}

\bibliographystyle{abbrv}

\end{document}